\documentclass[letterpaper, 10 pt, conference]{ieeeconf}
\IEEEoverridecommandlockouts                              
\usepackage[T1]{fontenc}
\usepackage{adjustbox}
\usepackage{formatting}
\newcommand{\figref}[1]{Fig.~\ref{#1}}
\newcommand{\secref}[1]{Section \ref{#1}}
\title{\LARGE \bf
Considerations for the Control Design of Augmentative Robots
}
\author{Shivani Guptasarma and Monroe Kennedy III
\thanks{The authors are with the Department of Mechanical Engineering, Stanford University, California 94305, USA {\tt\small \{shivanig, monroek\}@stanford.edu}}%
}
\begin{document}
\maketitle
\thispagestyle{empty}
\pagestyle{empty}
\begin{abstract}
Robotic systems that are intended to augment human capabilities commonly require the use of semi-autonomous control and artificial sensing, while at the same time aiming to empower the user to make decisions and take actions. This work identifies principles and techniques from the literature that can help to resolve this apparent contradiction. It is postulated that augmentative robots must function as \emph{tools} that have partial agency, as \emph{collaborative agents} that provide conditional transparency, and ideally, serve as extensions of the human body. 
\end{abstract}
\section{INTRODUCTION}\label{sc:intro}
As robotic systems advance in physical and computational capability, they become capable of both performing some tasks independently of humans, and performing other tasks alongside humans. However, when a robot is said to have been designed to augment human capabilities, it acts not alongside, but \emph{on behalf of} the human. Examples of such systems include powered prosthetic limbs, active exoskeletons, powered wheelchairs, wheelchair-mounted manipulator arms, semi-autonomous vehicles, teleoperated medical or field robots, and supernumerary robotic limbs. \\
In the IEEE Standard Ontologies for Robotics and Automation~\cite{ieeedefs}, a robot is defined as an \emph{agentive device}, where
\begin{itemize}
    \item an agent is ``something or someone that can act on its own and produce changes in the world", and
    \item a device is ``an artifact whose purpose is to serve as an instrument in a specific subclass of a process".
\end{itemize}
By this definition, \emph{any} robot has (a) the goal of behaving as intended (by human designers, programmers or users), and (b) a capability to make decisions to achieve this goal. An instrument or ``tool" satisfies only (a), behaving as intended when used by an expert. On the other hand, a fully autonomous robot is capable of decision-making (\figref{1a}), while it is also expected to achieve goals specified by its designers and programmers. If such a robot is assistive or collaborative in nature, it may be programmed in a manner that requires it to conform to the expectations of a human in the environment, yet its decisions are merely informed -- not controlled -- by human input. Augmentative robots, on the other hand, are distinguished from autonomous robots by the level at which human involvement occurs. There is a human \emph{operator}, user, or wearer, whose intention must be executed by the robot in a manner that the human experiences a sense of making their own decisions, acting and perceiving through the robot (\figref{1b}). In this way, using a good augmentative robot \emph{feels} like using a tool.
\subsection{Criteria for a robot to be an augmentative robot}
\begin{figure}[thpb]
      \centering
      \subfloat[\label{1a}]{
      \includegraphics[scale=3.3]{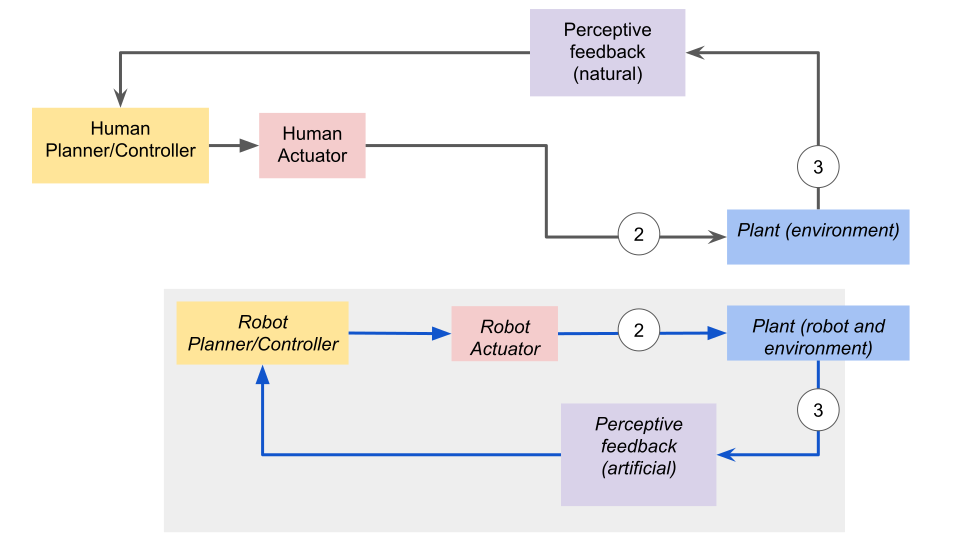}}
      \hfill
      \subfloat[\label{1b}]{%
      \includegraphics[scale=3.3]{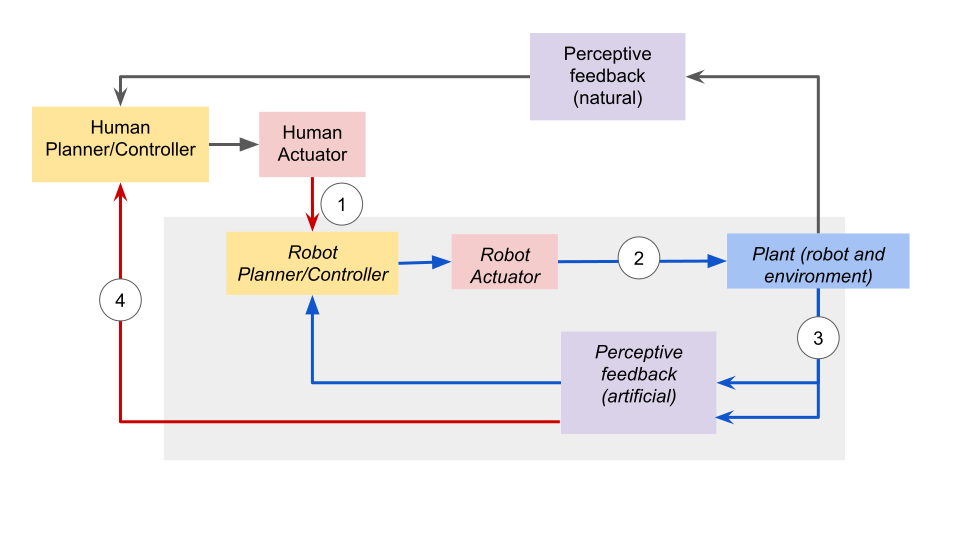}}
      \caption{(\figref{1a}) A human, without augmentation (top), and an autonomous robot (bottom); simplistically, the human planner/controller is the brain and spinal cord, while actuators are the muscles (\figref{1b}) A robotic system with human-in-the-loop control, providing an interface between a human operator and the environment (the encircled numbers represent the four functions described in \secref{sc:intro})}
      \label{fg:ctrldiag}
\end{figure}

An augmentative robot has four main functions, illustrated in \figref{fg:ctrldiag}:
\begin{enumerate}
    \item HUMAN INPUT: Acquiring control input from the human,
    \item ACTION: Performing actions in the environment,
    \item PERCEPTION: Perceiving the environment, and
    \item FEEDBACK: Conveying sensory feedback to the human.
\end{enumerate}
For the purpose of the current discussion, any robot that is required to perform (1) and (2) in such a manner that the human feels in control, and to perform (3) and (4) in such a manner that the human feels like the perceiver, is an augmentative robot. As is pointed out in~\cite{bergamascoHumanRobotAugmentation2016}, many of these are wearable robots. Additionally, teleoperated robots, mobility devices for manipulation and locomotion, and transport vehicles are included within the scope of this paper.
Section~\ref{sc:types} describes a classification of augmentative robots based on the nature and context of the augmentation provided by them; for general statements made in this paper, an effort is made to use examples from each of the two classes.
\subsection{Criteria for a device to be a robot}
Given that augmentative robots are required to provide the experience of using a tool, it is reasonable that they are designed for situations in which non-autonomous tools are insufficiently versatile or cumbersome to control. In this discussion, the presence of programmable computation in the control loop is sufficient to classify a device as a robot rather than a tool. For example, a prosthetic hook is a device that is a tool, whereas even a primitive myoelectric prosthesis is a robot, as it measures and interprets electromyographic signals to control motors. Similarly, a bicycle is a tool, while a vehicle with an engine management system (even if driven manually) is, in this sense, a robot, as is a semi-autonomously driven vehicle. Many factors determine whether a user prefers to augment their capabilities using a tool or using a robot; if there is a preference for tools even when the human's capabilities are severely limited, it is frequently related to shortcomings in the control design of augmentative robots (for example, among upper-limb prosthesis users, as the site of limb loss shifts from the elbow to the shoulder, the preference for passive devices increases, largely because myoelectric prostheses are difficult to control~\cite{chadwellRealityMyoelectricProstheses2016, markovicSensorFusionClosedloop2016}). Therefore, in \secref{sc:rto}, it is emphasized that robots, like tools, must provide \emph{transparency} in order to successfully augment human capabilities. \\
Finally, \secref{sc:principles} elaborates upon the principle of transparency in motion control and sensory feedback, mentions examples from the literature of ways to achieve transparency, and comments upon circumstances under which it may be desirable to reduce transparency in either control or perception.
\section{AUGMENTATION FOR LIFESTYLES OR PROCEDURES}\label{sc:types}
Certain augmentative robots, particularly those used by persons with disabilities, have been classified as ``assistive" technology in the literature, while others have been described as ``enhancing" existing capabilities~\footnote{In the current article, to avoid ambiguity, the use of the term ``assistive" is reserved for a robot that is perceived as an independent helper; for example, an autonomous care robot. Prostheses, exoskeletons, wheelchairs and wheelchair-mounted robotic arms are termed augmentative.}. It is of interest to consider the similarities and distinctions between these ``ability-substituting" and ``ability-extending" robots. 
\subsection{Common features of all augmentative robots}
Among the two categories of robotic systems described above, each is designed to interact with humans with physical limitations, and utilize particular kinds of inputs to help them overcome these limitations. In many applications, ranging from surgical micromanipulators to myoelectric prostheses, the robot must perform complex actions based on a limited number and quality of control inputs. In both ``ability-substituting" and ``ability-extending" robots, the user requires complex feedback information in order to adjust their control inputs, and this perception often needs to be provided to the user by the robot.
\subsection{Demands of the environment}
The difference between the two categories arises from the environment and context of use. Robots that are seen as ability-substituting are those which provide capabilities that the built environment is frequently not accommodating of. For example, a room may not be designed to be navigated in a wheelchair, whereas a road is designed to be navigated by a vehicle. As robotic surgery becomes commonplace, operation theaters may be designed to accommodate it; on the other hand, a person using a prosthetic hand may continue to encounter everyday objects that are not designed for easy manipulation.\\ 
As a result, robotic systems which are used continuously to interact with the built environment become part of the user's lifestyle. It is important to recognize that embodiment, agency, and psychological well-being are greater priorities in the design of such systems; whereas, in a robot designed to be used as part of a professional or short-term procedure, such a surgery or driving, it might be sufficient to guarantee safety and prioritize performance. 
\subsection{Demands of the user}
Another possible difference between the two categories of augmentative robots arises from the fact that individual human capabilities change over time. A robotic system designed for a person with an acquired disability may need to fulfill expectations that do not exist for a device providing an ability that the user has never experienced. Surveys of upper-limb prosthesis users have found explicit differences between user expectations and preferences based on whether the limb difference was acquired. For instance, in~\cite{biddissConsumerDesignPriorities2007}, users with acquired limb absence assigned greater value to shoulder and elbow movement, one of them mentioning that ``Without shoulder reach function, the prosthesis is more hindrance than help." A similar consideration applies to the design of devices for older adults.
\subsection{Overlaps between categories}
Certain devices may lie within either category based on context. For example, when an adult with congenital limb difference, who is already skilled in using their biological limb, decides to try on a prosthesis for the first time, their expectations might be qualitatively similar to a driver who has never used an autonomous vehicle, especially if they choose to use the limb for only certain specialized tasks (such as exercise or recreation). On the other hand, a prosthetic limb designed for day-long usage by a person with an acquired limb difference, who expects it to replace the functions of their biological limb, must be designed as a part of their lifestyle and with cognizance of the fact that it is replacing an ability demanded by either the user's previous experience, or the constraints of the built environment, or both. Similarly, an exoskeleton may be used for everyday ambulation or mountain-climbing; the differences in the choices available to the user govern the sequence of priority given to various design objectives. A recent study~\cite{vandijsseldonkExoskeletonHomeCommunity2020} reported that exoskeleton users with complete spinal cord injury find the device beneficial for community interactions, but relatively difficult to use in daily activities at home. In order of importance, ease of use, effectiveness, safety, and weight were listed by these users as the most important features.\\
Therefore, the two categories of augmentative robots may be referred to as ``lifestyle"-augmenting robots and ``procedure"-augmenting robots, to acknowledge a need to consider the differences in their usage and user expectations. While priorities in design depend upon whether a robot augments lifestyle or procedures, principles improving the control design of one category might still be valuable to the other, as performance, comfort, agency and embodiment cannot be dissociated with each other in a system that communicates intimately with a human user. For similar reasons, principles of assistive robotics, too, are often relevant to the development of augmentative robots, since identifying the needs and intentions of the human being being assisted is essential to the satisfactory performance of both.
\section{AUGMENTATIVE ROBOTS AS TOOLS AND ORGANS}\label{sc:rto}
\begin{figure}[thpb]
      \centering
      \includegraphics[scale=3.3]{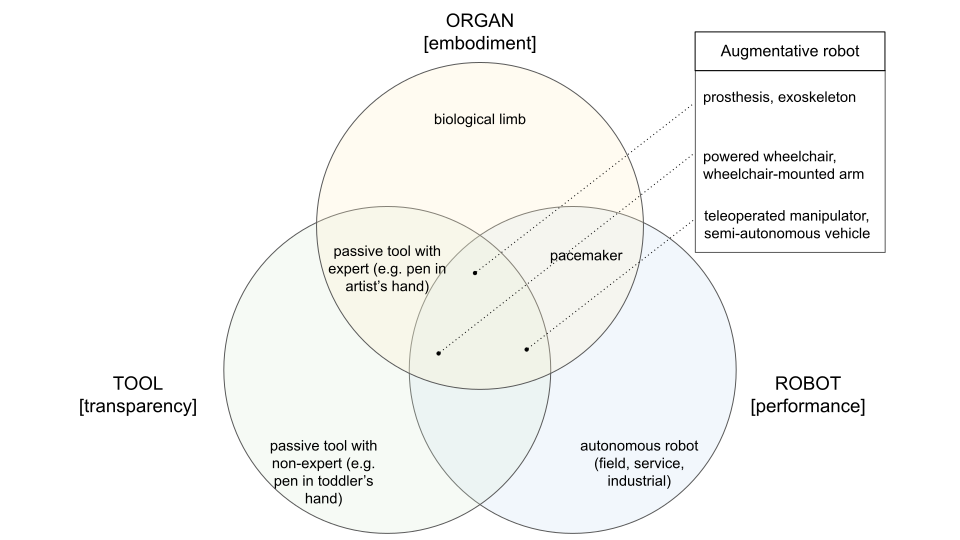}
      \caption{Augmentative robots lie at the intersection of robot, tool and organ, the relative emphasis on each aspect varying with application}
      \label{fg:venndiag}
   \end{figure}

Of the four functions described in \figref{fg:ctrldiag}, action and perception are common to autonomous robots and augmentative ones, while accepting control input and providing feedback, i.e., interfacing with and being integrated into the sensorimotor control of the human, are challenges unique to augmentative robots. While collaborative and assistive robots may interact with humans in their environment, it is only augmentative robots that act as extensions of the body and will of a single human user.\\
Therefore, as discussed in \secref{sc:intro}, an ideal augmentative robot functions as a tool, regardless of its degree of sophistication and even if it is partially automated. Unlike traditional, specialized tools, an augmentative robot is intended to be used for a wide variety of tasks; but, like all other tools, its purpose is to extend human capability by providing an effective means for the human to interact with the environment. In perceiving and acting upon the world, an augmentative robot may employ the same principles as its autonomous counterparts; yet, it may not assume agency. Instead, it is the human that acts, \emph{through} the robot. \\
\emph{Preserving and reinforcing a sense of agency (SoA) in the human user, therefore, is a central objective in the design of augmentative robots}. According to~\cite{mooreWhatSenseAgency2016}, SoA is defined as the ``feeling of being in the driving
seat when it comes to our actions". An augmentative robot which has a particular capability cannot be said to have succeeded in ``providing" that capability to its user, unless the user feels a strong SoA over the robot and its actions in the environment. The notion of SoA has proved challenging to explain, quantify and measure, although its implications for medical science and human-computer interaction have been recognized. A review of the methods of evaluating SoA in~\cite{mooreWhatSenseAgency2016} lists implicit as well as explicit measures of SoA. Implicit measurement techniques do not require the user to consciously consider whether they felt an SoA, and are arguably more reliable than surveys and questionnaires, as they instead measure a quantity correlated with voluntary action. Two of the most prevalent implicit measures are intentional binding and sensory attenuation. Intentional binding~\cite{haggard2002voluntary} is the phenomenon of an action and its effect being perceived as closer to each other in time, when the action is voluntary. Sensory attenuation~\cite{blakemore1998central} is a perception of reduced intensity for the effects of voluntary actions, relative to the effects of events not caused by the subject. Such effects are measured in specific experimental conditions, by assigning tasks such as pressing a key or touching a pad; methods which are difficult to adapt for complex and lengthy manipulation or locomotion tasks. As a result, robotics research is forced to rely upon qualitative and consciously-provided responses (the participant's impression of whether, and to what extent, they felt an SoA), potentially riddled with cognitive bias, in order to judge whether a device provides SoA.\textit{ Quantitative evaluation continues to depend heavily upon metrics of performance, such as speed and accuracy.}\\
Tool use typically extends both a user's sense of agency~\cite{haggardSenseAgencyHuman2017} and their neural representation of their body~\cite{maravitaToolsBodySchema2004}. The challenge in achieving SoA in augmentative robots arises from their dissimilarities with traditional tools, which are examined below:
\begin{enumerate}
    \item \label{multiskilled} A traditional tool (e.g., a screwdriver, or a spoon) is specialized for a range of functions, whereas an augmentative robot (e.g., a prosthetic arm or surgical teleoperated robot) \textit{needs to be capable of performing a wider variety of tasks}. 
    \item \label{duration} Accordingly, a traditional tool is intended to be used in a specific circumstance (e.g., while removing a screw, or eating/stirring), while an augmentative robot is \textit{used for an extended period of time} (e.g., most waking hours, or the duration of a surgery). Some tools -- such as hooks and other passive prostheses -- which are designed to be as versatile as possible, nevertheless cannot achieve the same level of general-purpose use as would an \emph{ideal} augmentative robot in the same situation.
    \item Following from~(\ref{multiskilled}), it is more likely for a traditional tool to have a similar level of complexity in its input (control) and output (action), whereas an augmentative robot \textit{needs to interpret low-dimensional and/or noisy input signals to perform complex and/or high-degree-of-freedom actions}. Sensory feedback to the human operator may also be severely limited when the task is complex. 
    \item \label{mismatch} As a result of the aforementioned mismatch in complexity between input control and output tasks, \textit{partial (or shared) autonomy is often an indispensable component} of the smooth control of augmentative robots. This in turn introduces an inherent risk of diminishing the user's sense of control~\cite{berberianAutomationTechnologySense2012, ciardoAttributionIntentionalAgency2020}.
    \item Specialized tools are typically treated, by a skilled user, as a natural extension of the human body, i.e., they provide a \emph{sense of embodiment}~(SoE). In the case of augmentative robots, however -- due to the extended duration of their use, the need to control complex tasks with simple inputs, and the possible presence of partial autonomy -- the \emph{difficulty} of ensuring a SoE, and the \emph{need} to ensure the same, are both heightened. SoE is distinct from SoA, as shown in experiments reviewed in~\cite{mooreWhatSenseAgency2016}, and is arguably even more elusive to measure, with most experiments being based upon the \emph{rubber hand illusion}~\cite{longoWhatEmbodimentPsychometric2008}, rather than a metric that could be generalized to devices with various functions and morphologies. Incidentally, while embodiment may be more crucial for the acceptance of lifestyle-augmenting robots than procedure-augmenting ones, it has been proposed (and stands to reason) that it would even improve the performance of the latter, e.g., in teleoperation~\cite{toetEnhancedTeleoperationEmbodiment2020}. It may be concluded that \textit{all augmentative robots need to provide some degree of SoE to the user}.
\end{enumerate}
To summarize, augmentative robots are (a) tools, in that they are intended to empower the user, (b) robots, in that they perform diverse tasks with partial autonomy, and (c) organs, in that a good tool operated by an expert user feels like a part of the body, and moreover, some augmentative robots are designed to imitate organs in form as well as function (some examples are shown in \figref{fg:venndiag}).
It follows that the philosophies of tool design and product design, robot decision-making and control, and the biological sciences, all have contributions to make to the problem of designing and controlling augmentative robots. 
\section{TRANSPARENCY IN THE CONTROL LOOP}
\label{sc:principles}
A biological organ containing voluntary muscles (limb, tongue, neck etc.) is connected to the decision-making brain by dense motor nerves and sensory nerves. The brain performs hierarchical control using the closed-loop formed by motor nerves, actuators (bones and muscles), sensors (primarily vision, hearing, touch, proprioception), and sensory nerves. \\
When the actuators and sensors in this loop are replaced by components of a robotic system, the control interface is often insufficient with respect to the complexity of the task; typical interfaces may consist of electromyographic electrodes, electroencephalographic electrodes, joysticks, sip-puff systems, manipulanda,  steering wheels, buttons, knobs and switches. Defining transparency as the \textbf{transmission of information without degradation}, in the forward direction of control, transparency implies that the user is (a) able to convey their intention to the robot and (b) predict the robot's response in order to perform an action in the environment. Only through transparency can the human develop \emph{trust} in the device: a belief that it has the same objective as themselves, as well as the ability to act towards that objective.
\subsection{Transparency in actuation}
Methods to improve transparency in actuation take the form of improved communication of human intent to the robot, and of robot intent to the human.
\begin{enumerate}
    \item \textit{Human-to-robot communication:} When direct input from the human is limited, estimating human intent using other means can greatly improve the performance and ease of use of augmentative robots. Recent research in prosthesis control has provided several inspiring examples of the same. Computer vision to identify objects of interest, followed by automatic grasp planning based on its shape and size, was shown to improve control of transradial prostheses in~\cite{markovicSensorFusionClosedloop2016}. Intuitive elbow movement was achieved in transhumeral prostheses in~\cite{meradCanWeAchieve2018} by using reaching movements performed with intact biological limbs to infer the natural relationships between shoulder and elbow joints.  In~\cite{zhuangSharedHumanRobot2019}, grasp stability with robotic hands was greatly improved by implementing local autonomy for control of finger joints. It is also possible to make use of other cues from the human, in addition to explicit instructions, as well as to design better ways of interpreting human input. Gaze tracking and electroencephalogram (EEG) input were together used to control manipulator arms in~\cite{zengSemiAutonomousRoboticArm2020}. In~\cite{junjeonSharedAutonomyLearned2020}, to make control of a~7-degrees-of-freedom~(DoF) robotic arm possible with a~2-DoF joystick, human control inputs are interpreted based on the confidence of the robot in its knowledge of the human's goal. The above examples make use of the following principles to improve the transparency of human intention to the robot:
    \begin{enumerate}
        \item gathering human input from multiple sources, actively and passively provided,
        \item strengthening perception and understanding of the environment by the robot, and
        \item increasing the complexity of processing algorithms for human input, based on generalizable principles.
    \end{enumerate}  
    \item \textit{Robot-to-human communication:} Making the quantities sensed by the robot, and the results of their processing, clearly and unobtrusively available to the human improves the human's internal model of the robot and, therefore, the quality of control input. In~\cite{schweisfurthElectrotactileEMGFeedback2016}, providing the human's own electromyographic control input as feedback was shown to cause an improvement in performance. Augmented reality is a powerful means to convey rich visual feedback to the user, and has been demonstrated to improve prosthesis control~\cite{clementeHumansCanIntegrate2017}, improve safety and quality of teleoperation in surgery~\cite{yamamotoAugmentedRealityHaptic2012}, and even make the use of robotic surrogates possible by people with severely limited mobility~\cite{griceInhomeRemoteUse2019}.
\end{enumerate}
In addition to estimating human intent and taking predictable actions to execute it, explicitly posing human empowerment as an objective of the robot also  contributes to transparency by bringing the behavior of the environment closer to the actions and expectations of the user. In the approach proposed in~\cite{duAvEAssistanceEmpowerment2020}, the robot takes actions which heighten the human's own ability to control the environment, having identified such actions through reinforcement learning. The results are demonstrated on a simplified task with shared autonomy; in this instance, the autonomous controller acts to stabilize a simulated lunar lander, making it easier for the human to control. \\
An effective augmentative robot must have transparency such that the control loop is unimpeded by the presence of the robot. This is achieved by combining the transparency of action with the transparency of perception, therefore allowing for fully transparent feedback control by the human.
\subsection{Transparency in perception}
Following the same definition of transparency as above, the requirement in the feedback direction is for the state of the environment to be conveyed to the human with as much fidelity as possible. Again, this may take various forms, e.g., a prosthetic limb faces challenges in conveying haptic feedback, although visual feedback is usually naturally available; a robot for minimally-invasive surgery needs to provide both visual and haptic feedback artificially. While using sensory information for closed-loop control within the robot helps the robot to achieve better task performance (e.g.,~\cite{zhuangSharedHumanRobot2019}), providing it back to the human can be an important component of both agency and embodiment. This is particularly true for feedback that the human is able and willing to make use of, to take decisions regarding actions, monitor performance, and adjust their control inputs to the robot. When it is possible to decouple the task into DoFs along which force or motion can be locally controlled and those which need to be controlled by the human, transparency in control can be achieved through haptic feedback of the same dimensionality as the control input, as has been described in~\cite{parkHapticTeleoperationApproach2006} in the context of teleoperation.\\
In situations where the human is either incapable of adjusting control inputs based on perception (reflex actions for safety), or does not wish to do so manually (mundane tasks), the device may take over control. These situations are discussed further in \secref{sc:ethics}. 
\subsection{Criteria for reduced transparency}\label{sc:ethics}
Transparency in action and perception demands that (a) the user can easily take the action that they wish to; and (b) they are aware of all aspects of the situation that might influence their decision. That is, when the user has both information and time to respond to the state of the environment, the augmentative robot must respond only in the manner dictated by the human. Thus, transparency of action and perception, if they can be achieved, transfer moral responsibility to the operator.\\
When either one of the criteria is not satisfied, there is a need for autonomous actions in accordance with ethical axioms. These actions provide an analogue to reflex actions in the human-robot control system. In a human body, reflex actions ensure that behaviors occuring on short time scales (too short for conscious consideration) are consistent with the individual's desire to preserve and protect the self and others from harm. Similarly, good augmentative devices are able to autonomously modulate their behavior to protect the operator from injury, whether continuously or in emergencies. A deliberate reduction of control transparency has been successfully demonstrated, for example, in~\cite{goilUsingMachineLearning2013}, where the the extent of control given to the user is modulated to blend it with autonomous control, to make it easier to avoid obstacles and navigate doorways in a powered wheelchair.\\
People also regularly make unconscious decisions to respect and preserve the lives and well-being of others; for example, avoiding pedestrians while driving a car, or avoiding a bystander's face while gesticulating in a conversation while wearing a prosthetic arm. When there is a risk of a behavior that might cause sudden harm, the device must be able to act autonomously, based on an assumption that the human operator is unable to perform the safe action, but upon conscious reflection, would have wished it to be performed. Transparency of action may be sacrificed in such situations based on the assumption that human intention is known and control inputs are unreliable if the human does not have time to respond. In~\cite{duanDriverBrakingBehavior2017},  the time taken by drivers to respond to impending collisions with cyclists was found to vary with their level of vigilance, which depended upon the direction of approach of the bicycle. Autonomous braking was proposed with adaptive timing so as to take safe actions without intruding upon the drivers' agency. The underlying assumption was that the driver does not wish to intentionally hit a cyclist. 
In teleoperated surgical robots, safety of the ``other" is inextricably linked with controllability for the user. It is standard, in such robots, to correct for tremor in the surgeon's hands: an obvious example of unreliable control input on a short time scale with the potential to cause harm, especially in regions such as the brain or the eye (see, e.g.,~\cite{roizenblattRobotassistedVitreoretinalSurgery2018}). \\
When the time scales involved in such decisions become larger and one may expect the human to respond, the role of the robot is to ensure that the human is provided with (a) an easy means to decide to take the safe action, and (b) a good understanding of the state of the environment, including factors that they may not have directly perceived. Warning systems in vehicles are examples of this strategy, where during reversing or changing lanes, the system alerts the user about nearby obstacles. Here, the assumption is that the user needed only the information that they might not have perceived; beyond being provided with the information, they are capable of -- and responsible for -- acting accordingly. Examples of such strategies abound in the literature on autonomous vehicles, but natural analogues can be seen in other areas where augmentative robots are only recently becoming commonplace, or where the degree of autonomy has only recently reached levels at which such questions become pertinent. For a semi-autonomous prosthesis or joystick-controlled arm that is capable of perceiving and interpreting the environment, ethical control at such time scales may take the form of setting relatively higher thresholds on human input signal for commanding movement towards obstacles, thus making it easy for the user to take safe actions, while ensuring that they retain the freedom to choose not to do so.\\
In addition to reflex actions, human organs perform another kind of unconscious task, namely, learned skills which have been committed to subconscious memory in the brain through repetition (some of which are colloquially termed ``muscle memory"). When performing repetitive tasks, the human may be capable of consciously directing the process, but may not wish to engage fully in it every time it is performed. Autonomous control is justifiable in this circumstance, however transparency must be readily accessible in case the human wishes to take conscious control. This would be classified as a ``takeover" in the terminology of~\cite{mccallTaxonomyAutonomousVehicle2019}, where the driver initiates a transfer of control from the vehicle. This principle may be expected to be central to improving the control of prostheses, where it is desirable to reduce cognitive load but imperative to allow the user to override any autonomous control. This is illustrated in~\cite{volkmarImprovingBimanualInteraction2019}, where both these objectives were achieved by automating the movement of one prosthetic hand in a bimanual task, while always providing the user with the ability to switch to myoelectric operation. \\
It is important to note that the preferences of users with regard to control method and degree of autonomy vary widely on an individual basis; studies such as~\cite{erdoganEffectRoboticWheelchair2017a, gopinathHumanintheLoopOptimizationShared2017} demonstrate this, respectively, for powered wheelchairs and joystick-controlled robotic arms with shared autonomy.\\
Finally, reduced transparency in perception may be desirable when the perceptive feedback would be detrimental to the operator's well-being or performance. For example, in a teleoperation scenario, auditory, visual and haptic feedback may be preferable to different extents during tasks requiring distributed or concentrated attention; a prosthetic hand may aim to convey information about pressure sensed in the most intuitive manner possible through applying pressure on a different part of the body, but it may not be desirable to convey high pressures beyond a limit where it is painful to the user.
\section{TOWARDS ORGAN-LIKE ROBOTIC TOOLS} \label{sc:conc}
Augmentative robots, as a general principle, must aim to achieve levels of transparency seen in traditional tools. Given the challenges of imperfect interfaces with the human operator, this can be achieved only by adopting strategies of autonomous control from traditional robots. They must also be easy to learn to use, allowing users to quickly become experts. Expert users would be able to perceive the tools as organs, i.e., extensions of the human body. Furthermore, augmentative robots must be able to perform reflex actions and unconscious tasks, similar to biological body parts, while ensuring that when the human has the time and desire to respond to changes in the environment, it is the human who bears the responsibility and freedom of action. In addition to quality of task performance and decision-making, both of which have advanced tremendously in the field of autonomous robotics, an emphasis on developing and using rigorous, quantitative metrics for agency and embodiment can lead to the design of augmentative robots that truly empower humans beyond their physical limitations. 
\section*{ACKNOWLEDGMENT}
The authors thank the anonymous reviewers for comments which improved the clarity of the paper. This work was supported by a Knight-Hennessy Scholars fellowship to the first author. 
\bibliographystyle{IEEEtran}
\bibliography{ethics_refs, prosthetics_refs}
\end{document}